\pdfoutput=1

\documentclass[11pt]{article}

\usepackage{emnlp2021}

\usepackage{times}
\usepackage{latexsym}
\usepackage{amsmath}
\usepackage{latexsym}
\usepackage{times}
\usepackage[greek,english]{babel}
\usepackage{latexsym}
\usepackage{hyperref}
\usepackage{caption}
\usepackage{xurl}
\usepackage{array}
\usepackage{graphicx}
\usepackage{xcolor}
\usepackage{color, colortbl}
\usepackage{multirow}
\usepackage{bm}
\usepackage{enumitem}
\usepackage{caption}
\usepackage{subcaption}
\usepackage{endnotes}
\usepackage[T1]{fontenc}

\usepackage[utf8]{inputenc}

\usepackage{microtype}

\newcounter{notecounter}
\newcommand{\enotesoff}{\long\gdef\enote##1##2{}}

\enotesoff

\title{BERT Cannot Align Characters}

\author{Antonis Maronikolakis \and 
	Philipp Dufter  \and
	Hinrich Sch\"{u}tze \\
	Center for Information and Language Processing (CIS), LMU Munich, Germany\\
	{\tt {antmarakis,philipp}@cis.lmu.de}}

\date{}

\begin{document}
\maketitle
\begin{abstract}

In previous work, it has been shown that BERT can adequately align cross-lingual sentences on the word level. Here we investigate whether BERT can also operate as a char-level aligner. The languages examined are English, Fake-English, German and Greek. We show that the closer two languages are, the better BERT can align them on the character level. BERT indeed works well in English to Fake-English alignment, but this does not generalize to natural languages to the same extent. Nevertheless, the proximity of two languages does seem to be a factor. English is more related to German than to Greek and this is reflected in how well BERT aligns them; English to German is better than English to Greek. We examine multiple setups and show that the similarity matrices for natural languages show weaker relations the further apart two languages are.

\end{abstract}

\section{Introduction}

For the many sweeping successes BERT has had in the field of Natural Language Processing, the model's alignment capabilities have been lacking and under-explored. Work in this area is picking up and it has been shown that BERT can operate with adequate efficiency in word alignment tasks \cite{attention_word_align,simalign}. The question whether or not BERT can perform character-level alignment, though, has not been answered yet.

Even though characters on their own do not necessarily hold much semantic meaning, we investigate whether BERT is able to generate useful representation spaces for characters. Character-level alignment would be useful in tasks like transliteration \cite{li-etal-2009-report,sajjad-etal-2017-statistical} or word-level alignment \cite{legrand-etal-2016-neural}. In a lot of occurrences of transliterations, grammatical inflections are added or dropped, which causes difficulties in an array of tasks \cite{czarnowska-etal-2019-dont,vania-lopez-2017-characters}. With character-level awareness, we can have better models for transliteration detection and extraction tasks. Word alignment could also benefit from character-level information in instances where words get split up within a sentence (eg., separable verbs in German or phrasal verbs in English).

With our work we show that even though BERT is not able to align languages on the character level, the closer these languages are the better the alignment. In the trivial case of English to Fake-English alignment, the model successfully learns to align characters. For English to German performance drops substantially and it drops even more for English to Greek. Languages seem to be put on an intuitive scale by BERT, with more similar languages having better alignment than highly dissimilar ones.

\begin{table}
	\centering
	\scriptsize
	\subfloat[$EngDeu$]{
		\begin{tabular}{|lll|}
			\hline
			\textbf{a} & \textbf{a} & 0.96\\
			b & z & 0.94\\
			c & l & 0.95\\
			d & x & 0.92\\
			e & h & 0.96\\
			\textbf{f} & \textbf{f} & 0.94\\
			\textbf{g} & \textbf{g} & 0.97\\
			h & x & 0.93\\
			\textbf{i} & \textbf{i} & 0.97\\
			j & z & 0.97\\
			k & c & 0.95\\
			l & z & 0.96\\
			\textbf{m} & \textbf{m} & 0.96\\
			\textbf{n} & \textbf{n} & 0.98\\
			\textbf{o} & \textbf{o} & 0.98\\
			\textbf{p} & \textbf{p} & 0.97\\
			q & c & 0.96\\
			\textbf{r} & \textbf{r} & 0.98\\
			s & x & 0.94\\
			t & k & 0.93\\
			u & z & 0.93\\
			v & h & 0.94\\
			\textbf{w} & \textbf{w} & 0.96\\
			\textbf{x} & \textbf{x} & 0.98\\
			y & c & 0.97\\
			z & c & 0.94\\
			\hline
	\end{tabular}}
	\quad
	\subfloat[$EngEll$]{
		\begin{tabular}{|lll|}
			\hline
			a & $\rho$ & 0.98\\
			b & $\xi$ & 0.98\\
			c & $\iota$ & 0.96\\
			d & $\sigma$ & 0.96\\
			e & o & 0.98\\
			f & $\iota$ & 0.98\\
			g & $\lambda$ & 0.98\\
			h & $\tau$ & 0.98\\
			i & $\tau$ & 0.98\\
			j & $\eta$ & 0.96\\
			k & $\nu$ & 0.97\\
			l & $\nu$ & 0.98\\
			m & $\psi$ & 0.96\\
			n & $\eta$ & 0.95\\
			o & $\alpha$ & 0.97\\
			\textbf{p} & \pmb{$\pi$} & 0.97\\
			q & $\rho$ & 0.97\\
			r & $\lambda$ & 0.99\\
			s & $\nu$ & 0.99\\
			\textbf{t} & \pmb{$\tau$} & 0.98\\
			u & $\phi$ & 0.97\\
			v & $\pi$ & 0.99\\
			w & $\tau$ & 0.98\\
			x & $\nu$ & 0.99\\
			y & $\nu$ & 0.98\\
			z & $\tau$ & 0.98\\
			\hline
	\end{tabular}}
	\caption{Max alignments for English $\rightarrow$ German and English $\rightarrow$ Greek, as extracted from a cosine similarity matrix between the two alphabets for each experiment. Correct alignments are in bold. We see low accuracy for both setups, but especially low for English $\rightarrow$ Greek.}
	\label{alignments_natural}
\end{table}

\section{Related Work}

Research has been conducted to uncover elements of multilinguality in mBERT. In \citet{how_multilingual_is_bert}, an analysis of mBERT is presented, while in \citet{cross_lingual_lm,wu-dredze-2019-beto,artetxe-etal-2020-cross} zero-shot cross-lingual transfer is analyzed. \citet{mbert_elements} further analyzes mBERT's capabilities with BERT's architecture and the structure of languages examined. The authors performed their experiments on a pairing of English with Fake-English, as proposed by \citet{crosslingual_study} in their rigorous empirical study of mBERT where linguistic properties of languages, architecture and learning objectives are investigated.

In \citet{char_level_transducer,garg-etal-2019-jointly}, it is shown that transformers \cite{transformers} can achieve similar performance with sequence-to-sequence approaches based on Recurrent Neural Networks \cite{luong-etal-2015-effective} for character-level tasks such as transliteration and grapheme-to-phoneme conversion. Further work to develop character-level BERT-based models is conducted in \citet{el-boukkouri-etal-2020-characterbert,ma-etal-2020-charbert}.

\citet{stat_alignment_models,legrand-etal-2016-neural} worked towards representation-based word alignment, with implementations of aligners proposed in \citet{fastalign,eflomal}.

We also note recent efforts towards unsupervised word alignment. In \citet{attention_word_align}, an extension to the usual Machine Translation encoder-decoder is proposed to jointly learn language translation and word alignment in an unsupervised manner. BERT has also been shown to be able to perform word alignment \cite{simalign} through embedding matrix similarities.

\section{Experimental Setup}

\subsection{Data Setup}

The EuroParl Corpus \cite{europarl} is a parallel corpus containing recorded proceedings of the European Parliament. Originally 21 languages were included, although here we examined three: English (ENG), German (DEU) and Greek (ELL).\footnote{We follow the ISO 639-3 standard for language codes: \url{https://iso639-3.sil.org/}} Each set is split by words and each word is further split in characters. Finally, special start and end tokens are added around each (split) word. This process results in a data file where each line contains a word split in characters. Finally, these language sets are merged together, alternating between lines. For example, in the English $\rightarrow$ German setup, the first line contains an English word (split in characters), the second line a German word (split in characters), and so on. For the conversion of English to Fake-English, we employ a mapping of characters to integers. The integers are in the range $[100, 151]$. The same mapping takes place for the English $\rightarrow$ German setup, since the two languages share the same script. Data is given as input line-by-line to the model.

\subsection{Model Setup}

We experimented with different BERT \cite{bert} model sizes and parameters. In our hyperparameter search, we mainly examined the effect of hidden layer size, layer numbers, embedding space size and attention heads number. We found that when models have fewer than 3 layers or more than 9 layers, we see underfitting and overfitting respectively. In the end, we settled for a 6-layer model with a quarter of the original BERT-base parameters, trained for 50 epochs. An analysis of model size effect on performance is omitted. We are training from scratch on the usual Masked Language Modeling task as described in \cite{bert}, with the difference that we are masking individual characters instead of subword tokens.

\subsection{Experiments}

For our control experiment, we tried to align English and Fake-English. This setup serves as aid to hyperparameter tuning. Because of the nature of English and Fake-English, if a model cannot align these two then it would not work for natural languages. English data was split in two sets, with the second converted to Fake-English. In our setup, Fake-English is a simple mapping from English characters to numbers ranging from 100 to 151. That is, `a' is converted to `100', `b' to `101', `A' to `126' etc. All other numbers were removed from both sets. We call this setup $EngFake_{base}$.

Apart from the base setup, we tried minor alterations. Namely, we tried to break the one-to-one mapping from English to Fake-English. The letter `f' was mapped not to a single, unique integer, but instead two new indices (denoted by $f_1$ and $f_2$). So, `f' was mapped to `200 201', which are the unique tokens for $f_1$ and $f_2$ respectively. In the same way, capital `F' was replaced by the tokens for $F_1$ and $F_2$. We call this setup $EngFake_{f_1f_2}$.

After the successful $EngFake$ experiments, we experiment with English to Greek ($EngEll$) and English to German ($EngDeu$). For English to German, since the languages share the same script, we converted German to Fake-German in an analogous method to the Fake-English conversion. Finally, German to Greek ($DeuEll$) experiments were also conducted for completion.

Firstly, we examine the uncontextualized embeddings of characters. To retrieve the representation of a character, we give it as input to the model and extract the embedding layer activations. We also investigate contextualized embeddings by feeding entire words into the model and extracting the activations for a particular character at the 5th layer.

For all the setups, a separate development set was used for evaluation by holding out 30\% of examples (ie., the lines in the dataset). After training the respective models, we compute the cosine similarity matrix between the two alphabets.

\section{Results}

\subsection{Similarity Matrix Comparisons}

The cosine similarity matrices are shown for our different setups. First, we give as input the characters of the two alphabets separately and extract their first-layer representations. We examined all layer representations, but since the characters are given without context, we decided to go with the first layer, which has been shown to contain context-independent information \cite{bert_probing}.

\begin{figure*}
    \centering
	\def\mywidth{0.75\textwidth}
	\def\mywidths{0.3\textwidth}
    \begin{subfigure}[t]{0.4\textwidth}
        \centering
        \includegraphics[scale=0.3125]{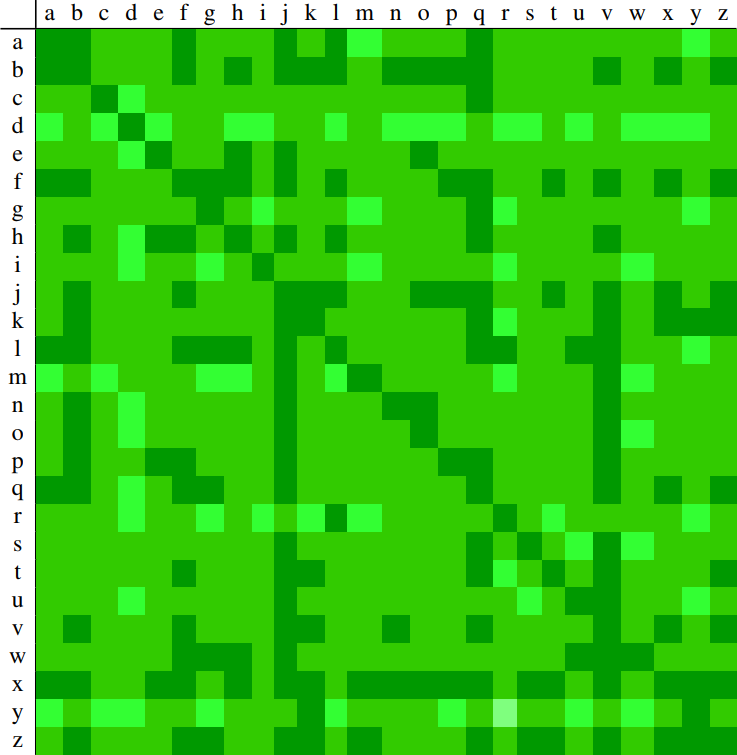}
        \caption{$EngFake$}
    \end{subfigure}
    ~ 
    \begin{subfigure}[t]{0.4\textwidth}
        \centering
        \includegraphics[scale=0.3125]{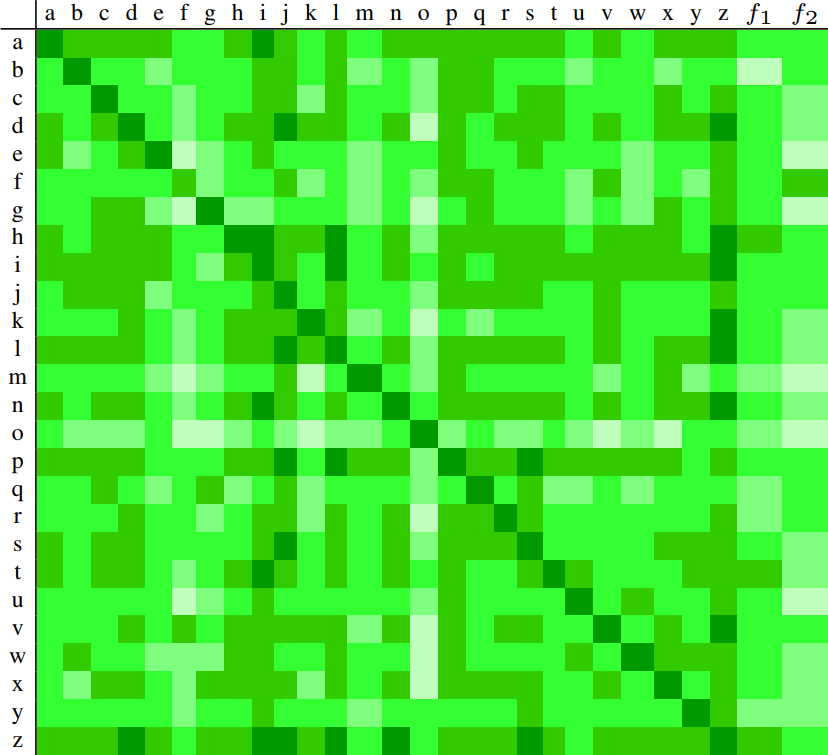}
        \caption{$EngFake_{f_1f_2}$}
    \end{subfigure}
    \caption{Showing heatmaps for (a) $EngFake$ and (b) $EngFake_{f_1f_2}$. The lighter green cells show lower cosine similarity, the darker green cells show higher cosine similarity.}
	\label{sim_en_fake}
\end{figure*}

In Figure \ref{sim_en_fake} (a) the cosine similarity matrix for $EngFake_{base}$ is presented. The diagonal shows that the model correctly aligns English with the base Fake-English language. In Figure \ref{sim_en_fake} (b) we see the similarity matrix of $EngFake_{f_1f_2}$. The strong diagonal indicates that BERT indeed manages to align English with its Fake-English equivalent. The added perturbation ($f_1f_2$ instead of `f') is correctly captured by the model as well. The English `f' has high similarity scores with both $f_1$ and $f_2$, with $f_1$ having a slightly higher score. We also compare against the combined $f_1f_2$ bigram, by computing its joint representation. We denote this new, combined token simply with `f' in the matrix.

\begin{figure*}
    \centering
	\def\mywidth{0.75\textwidth}
	\def\mywidths{0.3\textwidth}
    \begin{subfigure}[t]{0.5\textwidth}
        \centering
        \includegraphics[scale=0.35]{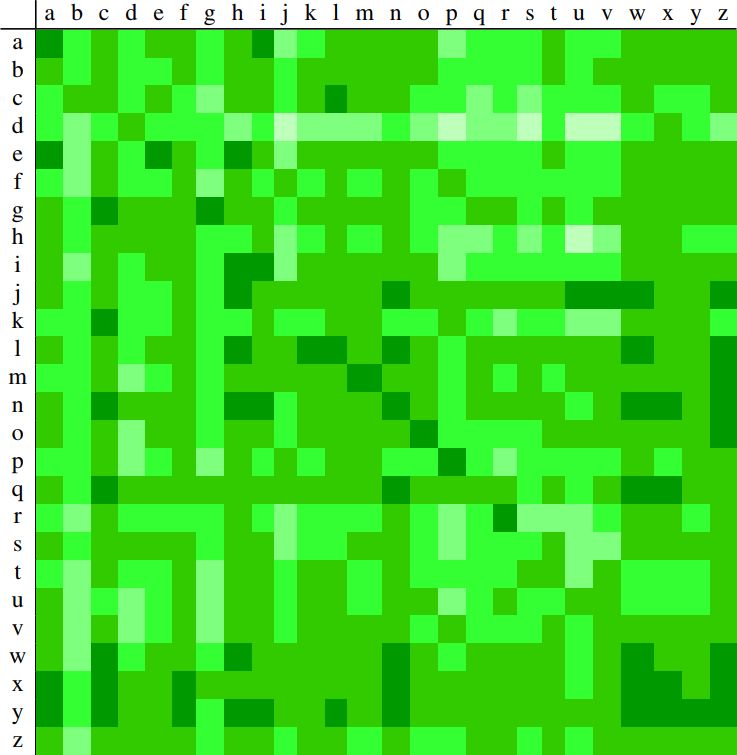}
        \caption{$EngDeu$}
    \end{subfigure}
    ~ 
    \begin{subfigure}[t]{0.5\textwidth}
        \centering
        \includegraphics[scale=0.35]{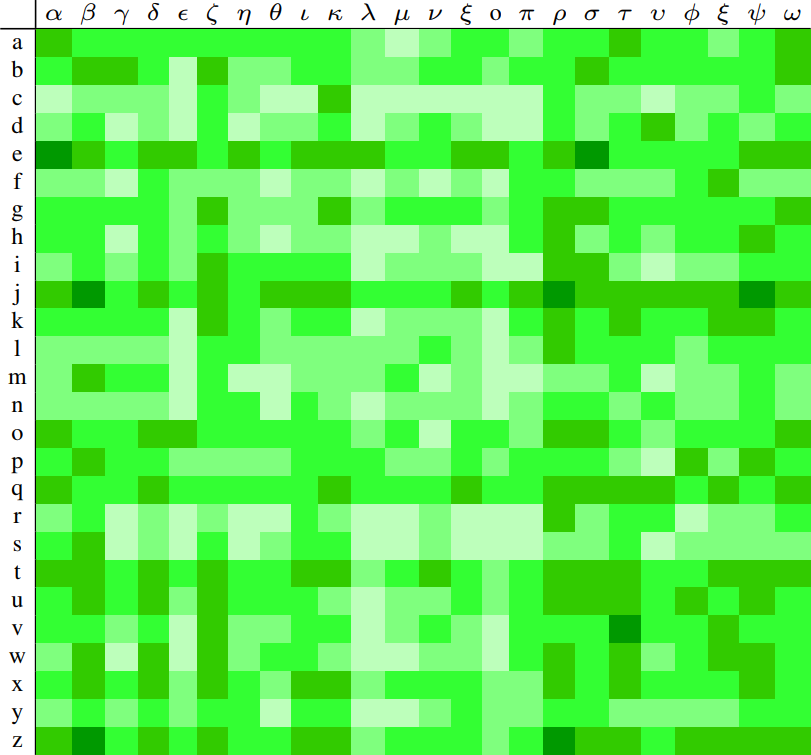}
        \caption{$EngEll$}
    \end{subfigure}%
    ~ 
    \begin{subfigure}[t]{0.5\textwidth}
        \centering
        \includegraphics[scale=0.35]{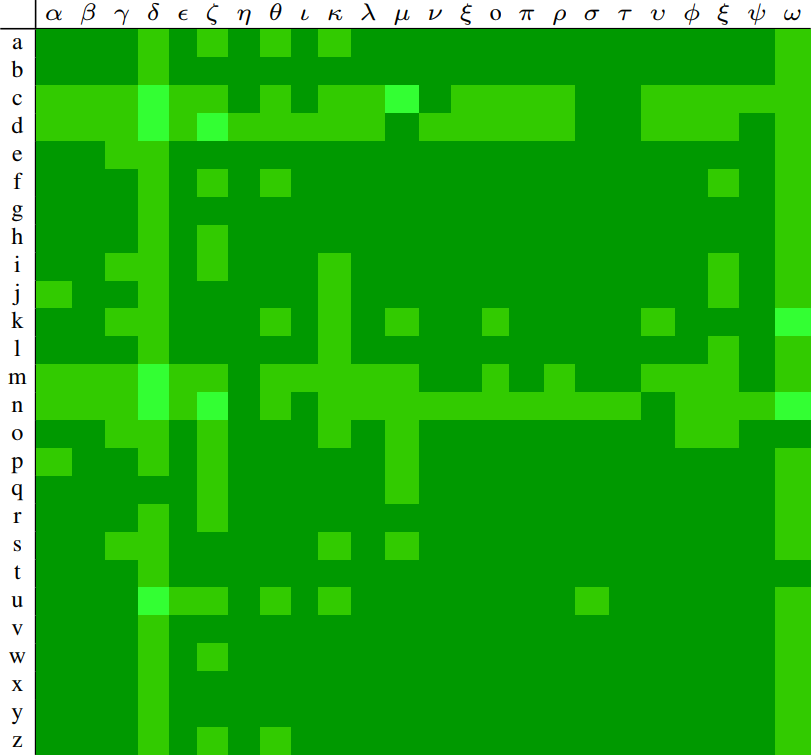}
        \caption{$DeuEll$}
    \end{subfigure}
    \caption{Showing heatmaps for (a) $EngDeu$, (b) $EngEll$ and (c) $DeuEll$. The lighter green cells show lower cosine similarity, the darker green cells show higher cosine similarity.}
	\label{sim_natural}
\end{figure*}

For $EngDeu$, in Table \ref{sim_natural} (a), we see lower performance. The diagonal is still observed, but less prominently than before. The two languages are relatively similar and belong to the same language family, so some similarity on the character level is to be expected. Nevertheless, the similarity matrix shows quite significant noise that obscures the diagonal. Note that even though the diagonal is not the best indicator of performance, the model was overall less efficient than expected.

One of the patterns that emerged is the high compatibility of the matrix's right hand side (German `w', `x', `y' and `z') with most of the English characters. High similarities can be found between these characters and most English characters across the board. This could be because of the low frequency of these letters in German.\footnote{\url{https://www.sttmedia.com/characterfrequency-german}} Apart from that, there are few other clusters around the matrix, for example `w', `x' and `y' for English and the `h'-`o' range in German.

When compared against a language further away from English than German, in this case Greek, similarities become even fainter. In Table \ref{sim_natural} (b) the similarity matrix for $EngEll$ is presented. We see weaker similarity scores across the board with results mostly random. As we show in Section \ref{max_align}, only a few characters were correctly aligned. In English, `e', `j', `t' and `z' have high similarity scores overall, with `$\rho$', `$\sigma$' and `$\tau$' in Greek scoring highly across the board as well.

Finally, in Table \ref{sim_natural} (c), $DeuEll$ is shown. There is very little that can be inferred from this matrix, since performance seems to be random.

\subsection{Max Alignment Accuracy}
\label{max_align}

Here we take a closer look at the similarity matrices and quantify how well our model aligns characters compared to a ground truth value. Even though ground truth alignment between characters is not always clear (for example, the English `a' has multiple pronunciations: `allure', `ball', `make', which would arguably map to three distinct characters in Greek), there are some obviously incorrect alignments we can observe (for example, the English `a' should never be aligned with any of the Greek consonants). Thus, this max alignment method is an adequate indicator of model performance.

In Table \ref{alignments_natural} we examine alignments for $EngDeu$ and $EngEll$. When choosing a target character, we search in the similarity matrix for the given pair of languages and choose the character with the maximum cosine similarity. In the natural language setups, the model fares fairly badly. In $EngDeu$ the model correctly aligns 11 characters (this is of course a mere simplification, since alignments such as `k' $\rightarrow$ `c' could also be considered correct; here we consider only the most basic of alignments: the ones on the diagonal). The situation is worse in $EngEll$, where only 3 characters were correctly aligned. $EngFake$ results are omitted, since the model correctly aligned all characters in all setups.

\subsection{Contextualized Embeddings}

We also perform a qualitative study on contextualized character-level representations. We choose three pairs of words, feed them separately to the model and extract their 5th-layer representations (averages of all layers were also examined with similar results, as well as other individual layers; the best performing layer was chosen). After computing the contextual representations for all characters in the English word, we align them with the contextual representations of the characters in the Greek word. In Table \ref{alignments_context} these alignments are shown.

Results are seemingly random with a strong bias towards the diagonal\footnote{To control for character positions, we also aligned characters without positional embeddings and results got worse.}. The model, thus, has not learned any contextual representations either.

\begin{table}
\centering
\footnotesize
\subfloat[monastery]{
\begin{tabular}{|lll|}
\hline
m & \textgreek{μ} & 0.63\\
o & \textgreek{ο} & 0.50\\
n & \textgreek{ν} & 0.37\\
a & \textgreek{α} & 0.49\\
s & \textgreek{η} & 0.23\\
t & \textgreek{τ} & 0.43\\
e & \textgreek{η} & 0.34\\
r & \textgreek{ρ} & 0.31\\
y & \textgreek{ι} & 0.22\\
\hline
\end{tabular}}
\quad
\subfloat[test]{
\begin{tabular}{|lll|}
\hline
t & \textgreek{τ} & 0.29\\
e & \textgreek{ε} & 0.21\\
s & \textgreek{τ} & 0.13\\
t & \textgreek{σ} & 0.34\\
\hline
\end{tabular}}
\quad
\subfloat[cardiac]{
\begin{tabular}{|lll|}
\hline
c & \textgreek{κ} & 0.77\\
a & \textgreek{α} & 0.63\\
r & \textgreek{ρ} & 0.55\\
d & \textgreek{ρ} & 0.19\\
i & \textgreek{ι} & 0.24\\
a & \textgreek{α} & 0.49\\
c & \textgreek{κ} & 0.52\\
\hline
\end{tabular}}
\caption{Max alignments for `monastery', `test' and `cardiac' as computed by our model.}
\label{alignments_context}
\end{table}

\section{Ablation Studies}

We also performed some minor experiments with static embeddings and adjustments to the MLM task, as well as a variation to our $EngFake$ setup.

Apart from the BERT embeddings, we also examined static embeddings, to introduce another baseline. Specifically, we used the FastText algorithm \cite{fasttext}. Data was the same as the previous experiments. Results were seemingly random, with FastText unable to capture any meaningful representations. Cosine similarities were generally very low and alignments incoherent.

For the MLM task, we experimented with different probabilities for token masking. In the original BERT paper, the chosen token is replaced by [MASK] 80\% of the time, while it remains the same 10\% of the time and gets replaced by a random token the rest 10\%. We experimented with the following distributions: 80/20/0, 60/20/20 and 50/50/0 with no noticeable change in performance.

We also experimented with overlapping Fake-English mappings in the form `a' $\rightarrow$ `100 101 102'. That is, each character is mapped to the tokens corresponding to itself and its next two characters (eg. `a' is mapped to the tokens for `a', `b' and `c'). This is an extreme case of the $EngFake_{f_1f_2}$ setup, where each character is mapped to three (instead of `f' getting mapped to `f1' and `f2'). Results deteriorated to random performance.

Finally, we conducted another $EngFake$ experiment. Namely, we restricted the context where we present each character. For the previous experiments each word was split in characters. Now, we further split each word into trigrams. So, instead of `e x a m p l e' we end up with multiple overlapping trigram entries: `e x a', `x a m', `a m p', etc. The motivation for this experiment is that since the transliteration of a character in a word doesn't need the entire word but instead only its direct neighbors, we should examine a setup with more restricted context\footnote{While this is a more nuanced subject, for the purposes of this ablation experiment we assumed this statement stood.}. In this case, results deteriorated heavily and are seemingly random with no meaningful representation captured.

\section{Conclusion}

In our work BERT is shown to be unable to create consistently good cross-lingual spaces on the character level. We train models on English, German, Greek and Fake-English and we compare character-level alignments between them. Cosine similarity matrices between the target and source alphabets were examined and we found that the closer two languages are, the better BERT does in aligning them. Fake-English is the easiest to align with English, whereas German is worse, with Greek trailing far behind. We conclude that BERT is not able to perform adequate character alignment.

\section*{Acknowledgments} This work was supported by ERCAdG \#740516. We want to thank the anonymous reviewers for their insightful comments and suggestions.

\newpage
\bibliography{anthology,custom}
\bibliographystyle{acl_natbib}

\end{document}